# Research Paper: Improving the Izhikevich Model Based on Rat Basolateral Amygdala and Hippocampus Neurons, and Recognizing Their Possible Firing Patterns

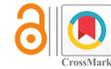

Sahar Hojjatinia[1] , Mahdi Aliyari Shoorehdeli[2] , Zahra Fatahi[3] , Zeinab Hojjatinia[4] , Abbas Haghparast[3]*

1. Department of Electrical Engineering and Computer Sciences, The Pennsylvania State University, Pennsylvania, USA.
2. Department of Electrical Engineering, K. N. Toosi University of Technology, Tehran, Iran.
3. Neuroscience Research Center, School of Medicine, Shahid Beheshti University of Medical Sciences, Tehran, Iran.
4. Department of Electrical Engineering, South Tehran Branch, Islamic Azad University, Tehran, Iran.

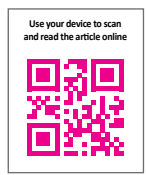

## ABSTRACT

**Introduction:** Identifying the potential firing patterns following different brain regions under normal and abnormal conditions increases our understanding of events at the level of neural interactions in the brain. Furthermore, it is important to be capable of modeling the potential neural activities to build precise artificial neural networks. The Izhikevich model is one of the simplest biologically-plausible models, i.e. capable of capturing most recognized firing patterns of neurons. This property makes the model efficient in simulating the large-scale networks of neurons. Improving the Izhikevich model for adapting with the neuronal activity of rat brain with great accuracy would make the model effective for future neural network implementations.

**Methods:** Data sampling from two brain regions, the HIP and BLA, was performed by the extracellular recordings of male Wistar rats, and spike sorting was conducted by Plexon offline sorter. Further analyses were performed through NeuroExplorer and MATLAB. To optimize the Izhikevich model parameters, a genetic algorithm was used. In this algorithm, optimization tools, like crossover and mutation, provide the basis for generating model parameters populations. The process of comparison in each iteration leads to the survival of better populations until achieving the optimum solution.

**Results:** In the present study, the possible firing patterns of the real single neurons of the HIP and BLA were identified. Additionally, an improved Izhikevich model was achieved. Accordingly, the real neuronal spiking pattern of these regions' neurons and the corresponding cases of the Izhikevich neuron spiking pattern were adjusted with great accuracy.

**Conclusion:** This study was conducted to elevate our knowledge of neural interactions in different structures of the brain and accelerate the quality of future large-scale neural networks simulations, as well as reducing the modeling complexity. This aim was achievable by performing the improved Izhikevich model, and inserting only the plausible firing patterns and eliminating unrealistic ones.

* **Corresponding Author:**
*Abbas Haghparast, PhD.*
*Address:* Neuroscience Research Center, Shahid Beheshti University of Medical Sciences, Tehran, Iran.
*Tel:* +98 (21) 22431624
*E-mail:* haghparast@yahoo.com, haghparast@sbmu.ac.ir

## Highlights

- We identified the possible firing patterns following the rat BLA and HIP neurons.

- We assigned a precise mathematical model to each recognized firing pattern of the BLA and HIP neurons.

- We explored the optimization parameters of the Izhikevich model based on the HIP and BLA data applying a genetic algorithm.

- We attempted to improve the Izhikevich model accuracy in representing rat data firing patterns.

- We defined the biological implication of changes in Izhikevich model values in optimization process.

## Plain Language Summary

One fundamental concept in understanding the functions of different brain areas, how they are connected to each other, and with which rate information transfers between the areas, is to recognize the firing pattern of neurons in those structures. These understandings promote the opportunity to successfully cure relevant neural diseases in the sense that the firing pattern of neurons in abnormal areas should be brought back to the normal firing activity. Additionally, identifying potential firing patterns for different structures of the brain and allocating a mathematical model to them reduces a great amount of complexity and implementation costs in the simulation of neural networks, based on the biological evidences. So, it is essential to identify the firing patterns that the neurons of different brain areas follow under normal and lesioned circumstances. Besides, it is important to develop possible treatments for neural diseases in a simulated sphere, as the first stage; the next step is to examine the effectiveness of them in a virtual condition before applying them to real-world (patients). To increase the accuracy of applied treatments, e.g., to return the abnormal firing activity of neurons to the normal one, the neural network should be generated based on the real data and performed by a biologically-plausible model. The Izhikevich model is among the best neural models that can represent different firing patterns with a fairly simple formulation. We identified the possible firing patterns following the HIP and BLA neurons as the primary goal. Second, to assign a mathematical model to each recognized firing pattern, the Izhikevich neural model was improved by optimizing its parameters for potential cases. Therefore, in the future simulation of neural networks, applying the improved Izhikevich model for the possible firing patterns could increase the accuracy of modeling and reduce its complexity. This achievement along with a greater understanding of firing activity of neurons under normal verses abnormal conditions results in more accurate simulated treatments based on the biological data; consequently, it leads to real world treatments in the future efforts.

## 1. Introduction

One of the pivotal components of the brain's microscopic structure is the neuron cell. The importance of this concept has led to extensive research studies to understand events at the level of individual neurons. One of the major findings was that unlike other body cells, neurons interact with each other by receiving and sending electric pulses or spikes.

Spiking Neural Networks (SNNs), i.e. the third neural networks generation (Maass, 1995), have been developed to imitate the natural neural networks. Spiking neural networks follow the same trend as computational neuroscience. The ultimate goal of both of them is to represent and configure the functionality of different brain areas realistically. The SNNs originated from the study of Hodgkin and Huxley (Hodgkin & Huxley, 1952), in 1952. The fundamental objective of SNNs is to encode the information of single spikes rather than just their firing rate (Maass & Bishop, 2001). Spiking neural networks have been used in numerous studies operating different applications. These approaches consist of regression and categorization (Dreiseitl & Ohno-Machado, 2002), deep learning (Sutskever, Vinyals, & Le, 2014), pattern recognition (Taigman, Yang, Ranzato, & Wolf, 2014), and behavioral prediction (Shen & Bax, 2013). Moreover, spiking neural networks facilitate the understanding of the human brain for researchers (Kuebler & Thivierge, 2014). Computational studies of SNN conducted by Maass and Schmitt (Maass, 1995; Maass, 1995a; Schmitt, 1998) have revealed the great efficiency of the third neural network generation.

In each dynamical study, one critical issue is, which model can describe the spiking dynamics of the neuron more efficiently. The Hodgkin-Huxley model can greatly stimulate the biological functioning of a neuron; however, it involves 12 equations consisting of 4 differential equations and 3 parameters to model one neuron's activity (Johnson & Chartier, 2017). This complex modeling results in a very expensive implementation. Additionally, the Hodgkin-Huxley model fails to exhibit the all-or-nothing firing mechanism for potential action generation (Deng, 2017).

A popular model that consists of a quite desirable compromise between computational efficiency and biologically-realistic behaviors is the Izhikevich model (Izhikevich, 2003). This model is not only biologically plausible, similar to Hodgkin-Huxley model, but also is computationally as efficient as an integrate-and-fire model. The Izhikevich model is also capable of simulating large-scale spiking neurons in real-time (Izhikevich, 2004). Therefore, to empower the neural interaction modeling based on real data, elevating the accuracy of the Izhikevich model in representing the neurons activity seems prominent. This will also result in increasing the application of Izhikevich model as an efficient model in implementing functional neural networks. One of the techniques to improve this model and adapt it to the behavior of considering real neurons is to optimize its parameters.

Genetic algorithms were developed according to basic concepts in the evolution and imitation of natural processes (Holland, 1975). These criteria consist of the mutation, recombination, and assortment of populations in a synthetic environment. The substantial components required in developing genetic algorithms were introduced by Bremermann in 1962 (Bremermann, 1962). Deciphering complicated problems by applying evolutionary techniques consisting of genetic algorithms, increased the popularity of these algorithms in the following years (Rechenberg, 1973; Schwefel, 1974). Genetic algorithm, i.e. one of the most popular evolutionary algorithms are applicable in solving optimization problems with a complex fitness landscape (Kellerer, Pferschy, & Pisinger, 2004; Conroy et al., 2019).

Two leading parts of the brain are the amygdala and the hippocampus. The amygdala is an influential area in memory pattern formation based on emotions (Tovote, Fadok, & Luthi, 2015; LeDoux, 2000), as well as the development of fear, anxiety, and corresponded diseases (LeDoux, 2000; Beyenburg, Mitchell, Schmidt, Elger, & Reuber, 2005). The amygdala plays a pivotal role in creating organisms' responses to their environment (Tovote, Fadok, & Luthi, 2015; Phelps & LeDoux, 2005). The hippocampus is among the major parts of the limbic system and an important area in strengthening memories, spatial learning, and emotional reactions (El-Falougy & Benuska, 2006). Therefore, the firing pattern identification of single neurons of these regions under the normal activity of the brain is of great importance. Furthermore, it is essential to be able to represent the firing activity of their neurons with a mathematical and biological model.

Developing an electrophysiological recording of single neurons activity provides a basis for exploring the structure of brain functions. However, the recorded signals are mostly contaminated by a high amount of background noise, noise from the recording system, or the activity of distant neurons. Moreover, the recorded data are related to the activity of several neurons adjacent to the recording site (Lewicki, 1998). Analyzing the massive amount of neural recordings requires one of the complicated interpretation tools, i.e. recognized as spike sorting. Spike sorting is the process of isolating action potentials from the background activity, i.e. considered as noise, extracting prominent spike features from the recognized spike waveforms, and finally allocating spikes with the same features to the neuron originated from that (Takekawa, Isomura, & Fukai, 2010; Rutishauser, Schuman, & Mamelak, 2006). This process can be conducted by an appropriate choice of clustering methods. In this paper, we used Plexon offline sorter software, i.e. a great and accurate tool for spike sorting.

Thus, one of the leading objectives of this study was to identify the possible firing patterns that the neurons of the HIP and BLA follow under normal activities. Another following noticeable purpose was to improve the Izhikevich model to make it more accurate in the sense of representing the firing activity of rat brain real data.

## 2. Methods

Male Wistar rats were used to investigate neuronal electric signaling in the normal BLA and HIP. Each rat was housed in the animal care facility maintained at 23±1○C and a 12:12 h light/dark cycle. Food and water were supplied to them with no limitation. The experimental processes were implemented based on the guidelines for the care and use of laboratory animals (National Institutes of Health Publication No. 80–23, revised 1996). All experiments were conducted in the Neuroscience Research

Center, Shahid Beheshti University of Medical Sciences, Tehran, Iran, according to the terms and conditions of the Research and Ethics Committee of this institute.

The study animals' anesthesia was achieved using urethane with an initial dose of 1.5 g/kg, Intraperitoneally (IP). Additional doses were given whenever required to maintain surgical anesthesia depth as checked by foot pinch and corneal reflex. To remove the potential pain, 0.1 mL buprenorphine was injected subcutaneously. Conducting tracheotomy, the study rats were located in a stereotaxic instrument. Using a heating pad, the rats' body temperature was maintained for the experiment duration. The electrophysiological recordings of the firing activity of neurons in the HIP and BLA were performed via an acute microelectrode with one channel. Each channel records the electric activity of a few neurons adjacent to it; the activity of farther neurons appeared as the background noise due to their low amplitude. The microelectrode proceeded to the left BLA (AP: -2.52 mm and ML: -4.8 mm from the bregma, and DV: -8.4 mm from the surface of skull) and the left HIP (AP: -3 mm and ML: -1.8 mm from the bregma, and DV: -3 mm from the skull surface) according to the rat brain atlas (Paxinos & Watson, 2007). Signals were recorded using a data acquisition system, filtered between 300 Hz and 10000 Hz, and sampled with a rate of 50 kHz. Each recording lasted 30 minutes.

The recorded data from the electric activity of neurons were exported to and analyzed via an offline sorter software, called Plexon (Plexon Inc., Dallas, TX). Spikes were detected through manual amplitude threshold discrimination. The threshold level discerns a trade-off between the missed spikes and the noise, which may pass that level. The threshold was assigned based on the amplitudes distribution of background activity and spikes. Next, spike sorting was performed to classify the electric activity of individual neurons, based on the first to third principle components, peak, valley, and other properties of signals. The Principle Component Analysis (PCA) is among the most effective linear spike feature extractors (Adamos, Kosmidis, & Theophilidis, 2008). Finally, spike clusters that represented a valid Inter Spike Interval (ISI) histogram (Theodoridis & Koutroumbas, 2009) were saved for further analysis. NeuroExplorer (Nex Technologies, Colorado Springs, CO) was used to analyze the firing activity of clusters of neurons. The quality of sorted data was validated through auto-correlogram analysis. Auto-correlogram displays a single spike train against itself. Another tool that compares the arrival times of spike trains is a cross-correlogram. Through cross-correlogram, the different identified clusters of spikes were explored to validate the exact number of neurons in each set of recorded data. Finally, the average firing rate histograms were generated and verified for all neurons, over the entire period of 30 minutes. Then, the validated clusters of spikes were exported to MATLAB to be used in modeling. This software was also used to code our desirable genetic algorithm, Izhikevich model, and represent the comparison figures of different firing patterns.

The two-dimensional Izhikevich neuronal model (Izhikevich, 2003) is defined by three Equations 1, 2 and 3, as follows:

1. $\dot{v}=0.04v^2+5v+140-u+I$

2. $\dot{u}=a(bv-u)$

3. $if\ v\geq+30\ mV, v\leftarrow c, u\leftarrow u+d$

Where variables and are the membrane potential of the neuron and membrane recovery variable, respectively. The activation of ionic currents and the inactivation of ionic currents can be represented by the variable . This variable supplies with a negative feedback. Variable represents the delivery of synaptic currents. Equation (3) activates when the amplitude of action potential reaches the threshold and are dimensionless parameters of the model.

Differences in the quantities of the Izhikevich model parameters result in the exhibition of various firing patterns that neurons may follow. The parameter traces the time scale of recovery variable . Therefore, smaller amounts of represent a slower recovery period. The parameter describes the sensitivity of variable to oscillations in membrane potential . Based on the values of this parameter, the resting potential is volatile between -70 and -60 mV (Izhikevich, 2007). The parameter indicates the after-spike reset value of the variable and has an amount between -50 and -65, in different patterns. Amount -65 determines deep voltage reset, amount -55 governs high voltage reset, and -50 represents moderate after-spike jump (Izhikevich, 2007). Parameter outlines the after-spike reset of the variable . This parameter changes in a wider range; its higher values reflect greater amounts of the after-spike jump of recovery variable . Figure 1 summarizes the mentioned explanations related to the parameters in a visual stand.

To simulate 1 ms of the Izhikevich model, the operation of only 13 floating points is required. This property makes the model highly effective in simulating large-scale networks of neurons (Izhikevich, 2003). Accord-ing to both points of view that are biological plausibility (number of features) and implementation cost (an approximate number of floating-point operations), the Izhikevich model is fairly in desirable condition to be used. Therefore, the model efficiency in representing the spiking behavior of rat brain neurons with great accuracy can fortify the model.

The Izhikevich model could exhibit all the firing patterns, i.e. shown in Figure 2. It illustrates various spiking patterns of individual neurons, based on their response to the applied Direct Current (DC) (Izhikevich, 2010). Some of the neuro-computational properties that the BLA and HIP neurons provide a similar neuronal behavior with them are also demonstrated in Figure 2. These properties are tonic spiking (Nessler, Pfeiffer, Buesing, & Maass, 2013), phasic spiking (Malsburg, 1999), mixed model (Connors & Gutnick, 1990), integrator, rebound spike, threshold variability (Izhikevich, 2003), depolarizing after-potentials (Malsburg, 1999), and inhibition-induced spiking (Izhikevich, 2003).

The genetic algorithm is among the well-known evolutionary algorithms that employ the principle of best populations' selection in each iteration for the whole process. This property provides the opportunity to select and generate individuals that are more adapted to the environment and remove the ones with less consistency. By repeating the same process for several generations and replacing undesirable populations with more adjustable ones, the algorithm evolves a population with optimal characteristics. The capability of a genetic algorithm in operating with continuous and discrete variables, as well as linear and nonlinear fitness functions, makes it a great candidate in solving complicated optimization problems (Hassan, Cohanim, & Weck, 2004). A basic genetic algorithm procedure consists of the following key components (Goldberg, 1989; Pelikan, 2010):

Initialization: Genetic algorithms produce the initial population of solutions arbitrarily. This generation conducts based on a unique distribution of admissible solutions. Selection: Over the course of each iteration, genetic algorithms select the more adjustable solutions from the existing set of populations. This process employs the more qualified solutions.

Variation: Two great tools recruited in genetic algorithms are crossover and mutations. Applying these tools to selected solutions in prior step, results in the generation of new solutions. Crossover is the process of recombining different subsections of promising solutions. Likewise, mutation applies instant alternation in integrated solutions.

Replacement: In this step, next-generation is produced by replacing the primary solutions or some parts of them with the new desirable ones, i.e. generated via crossover and mutation.

## 3. Results

As mentioned before, the parameters of the Izhikevich model have different values to exhibit different potential firing patterns of neurons. In this study, we optimized each set of parameters by modifying our optimization problem variables, such as maximum number of iterations, crossover percentage, mutation rate, etc., in the performed genetic algorithm to minimize the associated error.

In this paper, we recorded data from two regions of the rat brain consisting of the BLA and HIP under normal activity. By spike detection via Plexon Offline Sorter software, the data of each region were divided into 3 clusters. Based on spike sorting criteria, each cluster represents the activity of one single neuron adjacent to the recording site. Afterward, we compared the firing patterns of the original Izhikevich model, the model with optimized parameters and, and the firing behavior of the mentioned regions' real single neurons.

Initially, after spike sorting, we investigated each single neuron of the data in terms of following which of the firing patterns. Accordingly, we traced the activity of each neuron in a specific time period and compared them with the recognized firing patterns. The next step was to design a proper genetic algorithm to optimize the corresponding cases of the Izhikevich model parameters. As mentioned before, in designing the proper genetic algorithm, the values of optimization problem variables depend on different cases of the Izhikevich neural pattern and data. Various tests were run with different variable levels. The fitness criterion was the error minimization (mean square error) of the neural action potential difference between the Izhikevich and real neurons. As a case example, considering the designed genetic algorithm for tonic spiking pattern for the BLA neurons, the optimal crossover and mutation ratio were assigned 0.7 and 0.8 values, respectively. The algorithm terminated in 150 generations.

We represent the obtained results according to the potential firing patterns following the detected single neurons. Other cases of Izhikevich pattern that were excluded from further consideration and explanations in this study were as follows: the ones with considerable

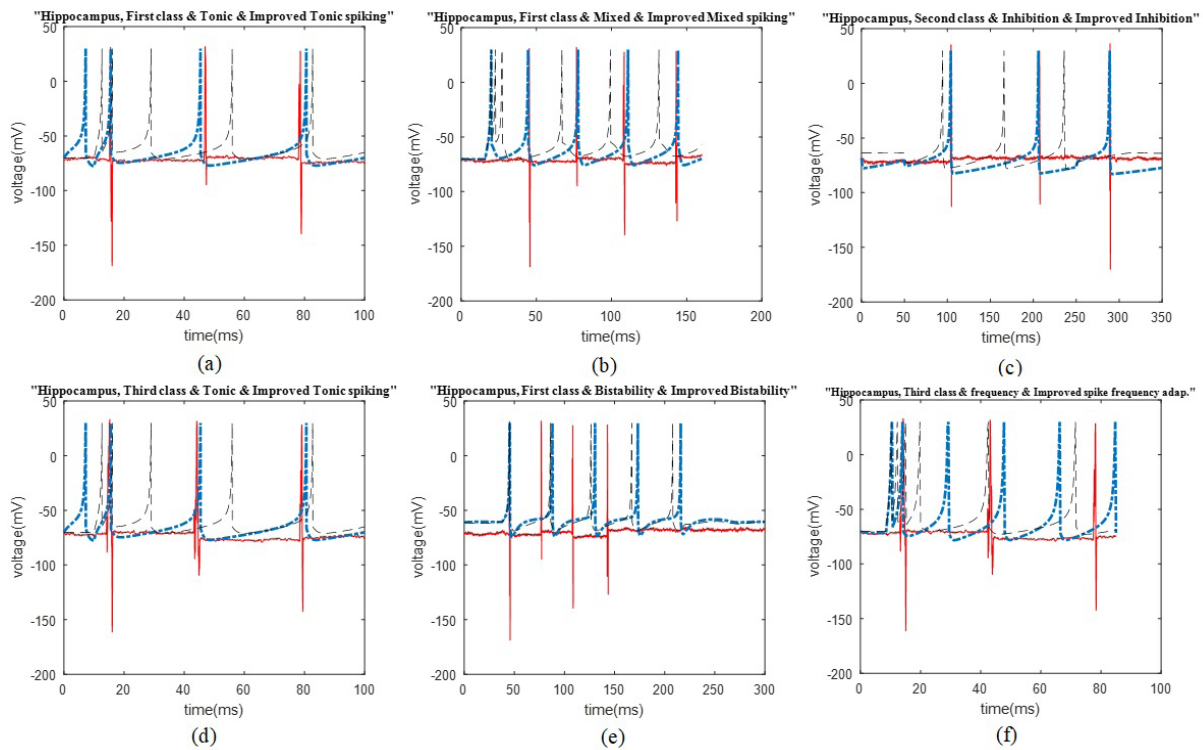

**Figure 1.** The BLA first cluster firing patterns comparison

This figure presents the comparison of the first cluster firing pattern of the rat BLA, possible firing patterns of the Izhikevich neuron, and the improved pattern, in distinct plots. The mentioned firing patterns are the integrator, phasic spiking, depolarizing after-potential, rebound spiking, and threshold variability.

**NEUR** SCIENCE

different spike timing or the number of spikes in a given time period. Some of these inconsistencies are shown in Figure 2 C-E, for the BLA, as well as Figure 3 E and F for the HIP.

We illustrated the comparison plots of possible firing patterns related to different clusters of the HIP and BLA, corresponding Izhikevich, and improved Izhikevich patterns in several figures. In all figures, the red/line curve is related to the study rats' real neuron spiking; the black/dash curve is relevant to Izhikevich neuron, and the blue/dash-dot curve is related to the improved Izhikevich neuron. Moreover, the study rats' real neuron spiking plot and Izhikevich neuron plot were not matched in most cases. Therefore, the Izhikevich model must be improved for adjustments. The initial jump in membrane potential represented in a few figures was not a spike; however, it is a transient mode in the neurons' firing activities.

The data recording and analysis were conducted on several rats and the results were desirably consistent. The outcomes of parameter optimization for the BLA single neurons in firing patterns are shown in Figures 1 and 2. According to the figures, the real neurons of the BLA greatly adapted with the improved Izhikevich neurons. Moreover, the achieved results represented that first region cluster may follow a firing pattern of each of the improved integrator, phasic spiking, depolarizing, rebound spiking, or threshold variability (Figure 1). The second and third clusters followed a firing pattern of the improved Izhikevich pattern for inhibition-induced spiking and tonic spiking, respectively. The activity pattern of these two clusters is presented in Figure 2 A and B.

In Figure 2 C-E, we showed three neural behaviors that a real neuron of the BLA was not followed. Then, we compared them with either the firing pattern of Izhikevich neuron or Izhikevich-improved neuron. The mentioned firing behaviors consisted of mixed spiking, bistability, and spike frequency adaptation.

Similar to the previous discussed region of the rat brain, the HIP neurons have greatly followed the improved Izhikevich pattern for each sorted cluster (Figure 3 A-D). For all considered rats, the three single neurons extracted from data recording of this region had the firing pattern, as follows: one cluster followed improved mixed or tonic spiking, the second cluster followed the

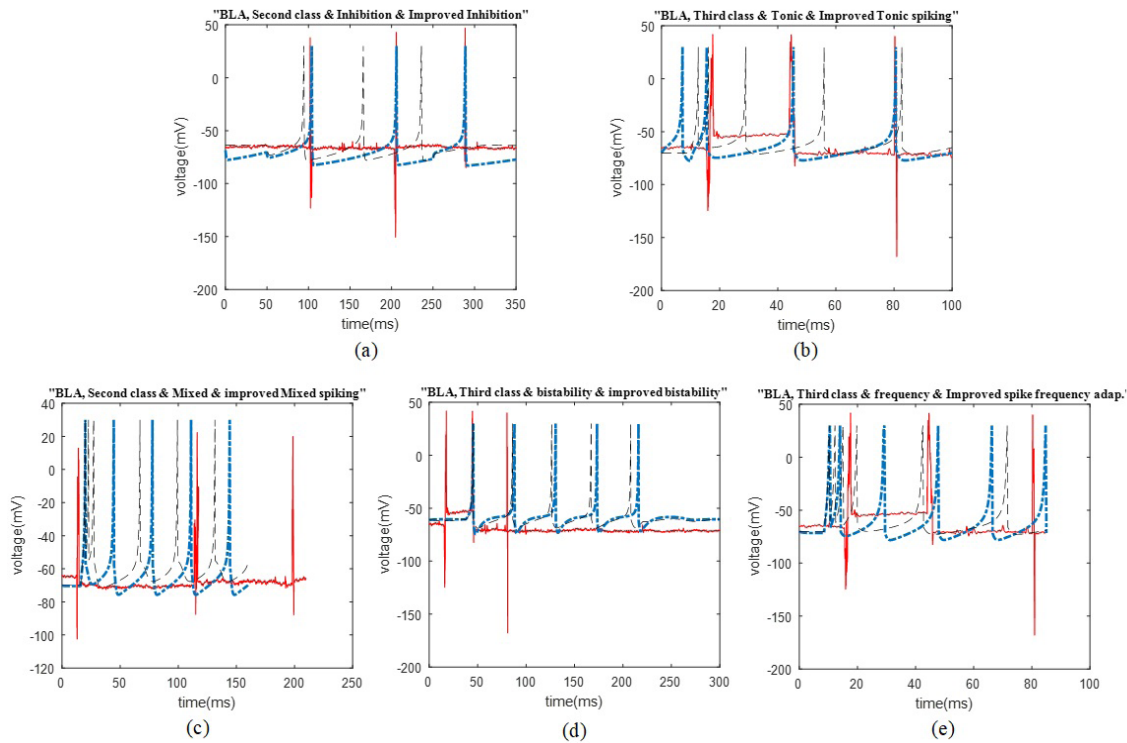

**Figure 2.** The possible firing patterns of the BLA neurons

NEUR SCIENCE

This figure represents the comparison of the second and third cluster firing patterns of the rat BLA, possible firing patterns of the Izhikevich neuron, and the improved patterns, in plots (a) and (b). The second cluster may follow inhibition induced spiking, and the third cluster may track tonic spiking. Additionally, three possible firing patterns that may not be followed by the BLA neurons are shown in plots (c) to (e). In these plots, the comparison of second and third cluster firing patterns of the BLA, mixed spiking, bistability, and spike frequency adaptation from the Izhikevich firing pattern and improved ones are illustrated.

improved inhibition-induced spiking, and the last one followed improved tonic spiking. Figure 3 E and F represent two firing activity which the HIP single neurons may not follow. These patterns are bistability and spike frequency adaptation.

## 4. Discussion

Several significant results were achieved based on the in vivo electrophysiological data in this study. First, the structure behind the firing patterns following the single neurons of the rat BLA and HIP were identified. This finding improves our understanding of the behavior of different structures in the nervous system. Second, a precise mathematical model was assigned to each recognized firing pattern. This aim was achieved using one of the most effective neuronal spiking models for large scale simulations; Izhikevich model, due to its great trade-off between simplicity, computational feasibility, and biological plausibility. To reach the proper adjustable mathematical model for the BLA and HIP neurons, the Izhikevich model parameters were optimized for all different possible cases, using a genetic algorithm. This achievement increased the Izhikevich model's accuracy in representing a mathematical model for the real neurons of rat brain. These findings greatly impact the future modeling of networks of neurons consisting of the BLA and HIP in which the unlikely firing patterns can be excluded from the consideration. The elimination of unrealistic firing patterns and performing the proper mathematical model for the probable ones will result in elevating the quality of neural network simulations and reducing the complexity of the modeling. Finally, the biological implication of changes in the Izhikevich model values in the optimization process was defined.

As mentioned earlier, the Izhikevich model is capable of representing the firing pattern of most recognized types of cortical neurons according to changes in the values of its parameters. However, it fails to represent the neuronal firing pattern of some specified parts of the rat cortex, such as the HIP and BLA, with great accuracy. As per the real data in this study, one potential problem of the Izhikevich model was its post spike potential; the Izhikevich neuron potential returns to the amount of parameter, (Figure 1) (Izhikevich, 2003). Nevertheless,

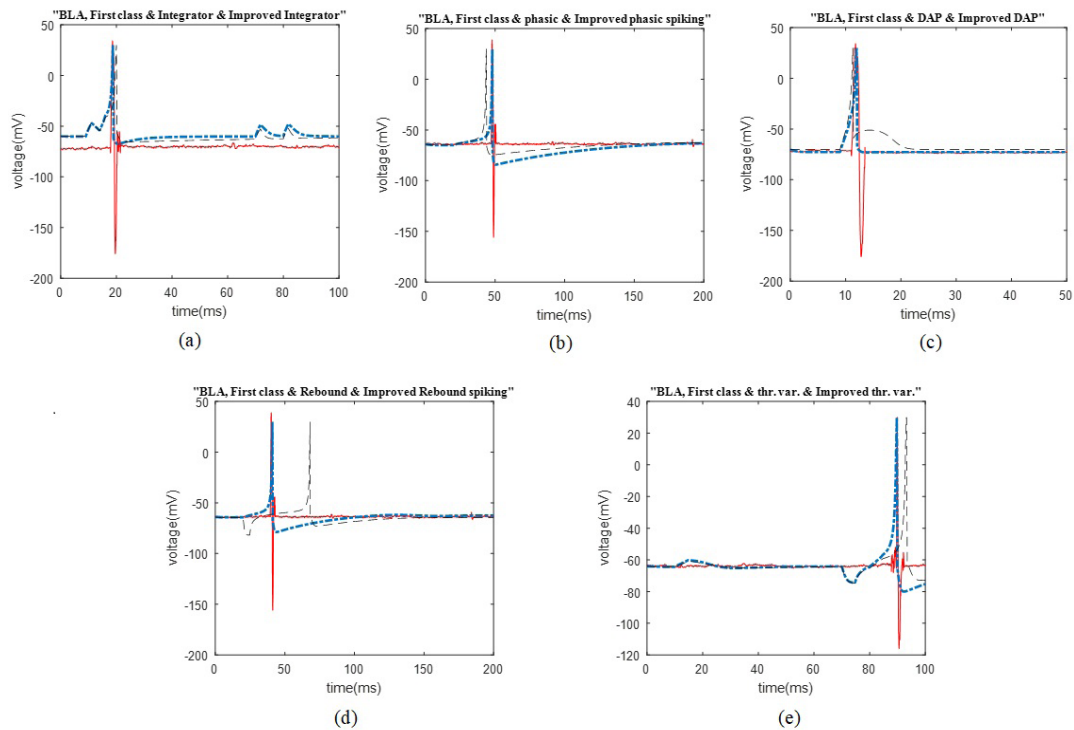

**Figure 3.** The comparison of the HIP neurons firing patterns

Figure 3 A-D shows the comparison of the possible firing patterns of the rat HIP that are tonic spiking, mixed spiking, and inhibition induced spiking mode from Izhikevich firing pattern, and improved patterns. Figure 3 E and F represent unfollowed firing patterns that are bistability for the first cluster and spike frequency adaptation for the third cluster of the HIP neuron.

NEUROSCIENCE

considered cases in current exploration did not follow this potential reset. They returned to their initial amount (Figures 1-3).

The Izhikevich model is a well-known spiking neuronal model used in numerous computational neuroscience investigations. Some studies that have applied the Izhikevich neural model are represented in the following. Zhao et al. investigated the probability of detecting a weak electric field in neural networks with the presence of noise (Zhao et al., 2017). Lv et al. have implemented the simultaneous simulations of brain networks based on the Izhikevich spiking model (Lv et al., 2014). Mizoguchi et al. have developed a silicon neuron circuit based on the Izhikevich neuron (Mizoguchi, Nagamatsu, Aihara, & Kohno, 2011). Nageswaran et al. have demonstrated an efficient, biologically realistic large-scale spiking neural model simulator that runs on a graphics processing unit in their research; the model included Izhikevich spiking neuron (Nageswaran, Dutt, Krichmar, Nicolau, & Veidenbaum, 2009).

As a result, the Izhikevich neural model has been used in various studies and is highly applicable in neural network simulations. The popularity of this model in large scale simulations is because of its simplicity in implementation, as well as its biological plausibility. Accordingly, improving the model by increasing its accuracy in adjustment with brain real data spiking activity could result in the outcome improvement of studies conducted based on this model. Improvements can be reflected in either approaching a more realistic result or designing a more reliable neural network. The necessity for model modification led us to investigate the potential techniques to improve the Izhikevich model. To achieve this purpose, one of the best methods represented in this paper was to optimize the Izhikevich model parameters. This enhancement enables the model to adjust with the rat cortex neuron spiking pattern with great accuracy. Besides, it promotes the efficiency of investigations based on the Izhikevich model.

Optimization is the process of improving a conception, based on the obtained information. In problem optimization, the goal is to achieve the best solution, even with the presence of a large amount of noise (Ravazzi et al., 2018). This concept declares that there exist several solutions to the problem; each has a different value

**Table 1.** Comparison of the prior and anterior parameter values of Izhikevich neural model for various firing patterns under optimization.

| variables | Status | Parameter | Parameter | Parameter | Parameter |
|---|---|---|---|---|---|
| Tonic spiking | Without optimization | 0.02 | 0.2 | -65 | 6 |
|  | With optimization | 0.01877 | 0.26801 | -66.3083 | 12.46620 |
| Mixed spiking | Without optimization | 0.02 | 0.2 | -55 | 4 |
|  | With optimization | 0.037413 | 0.19586 | -56.165 | 10.82459 |
| Integrator | Without optimization | 0.02 | -0.1 | -55 | 6 |
|  | With optimization | 0.1975 | -0.1025 | -59.874 | 12.16331 |
| Depolarizing | Without optimization | 1 | 0.2 | -60 | -21 |
|  | With optimization | 1.738625 | 0.165259 | -67.823 | -4.4429 |
| Phasic spiking | Without optimization | 0.02 | 0.25 | -65 | 6 |
|  | With optimization | 0.023135 | 0.24904 | -67.904 | 19.35829 |
| Rebound spiking | Without optimization | 0.03 | 0.25 | -60 | 4 |
|  | With optimization | 0.038877 | 0.25154 | -64.0123 | 11.09948 |
| Threshold variability | Without optimization | 0.03 | 0.25 | -60 | 4 |
|  | With optimization | 0.067733 | 0.251266 | -62.2565 | 13.03464 |
| Inhibition-induced spiking | Without optimization | -0.02 | -1 | -60 | 8 |
|  | With optimization | 0.0090563 | -1.14721 | -62.0312 | 16.2566 |

**NEUROSCIENCE**

(Haupt & Haupt, 2004). Solutions should be determined by considering the situation and conditions. Optimization algorithms divide into 6 categories; some of them aim to minimize the cost. Although the minimum seeker algorithms are usually fast, they fail to distinguish the local minimum solutions from the global ones. In contrast, optimization algorithms, like genetic algorithm, are more successful in achieving the global minimum while decreasing the processing speed (Haupt & Haupt, 2004).

Complex real-world problems and attempts to find appropriate solutions for them have led scientists to investigate natural phenomena and imitate them for years. Optimization algorithms have been progressively developed based on the natural processes in the past decades (Michalewiez, 1996). Some outstanding algorithms, such as evolutionary algorithms (Back & Schwefel, 1996) and the genetic algorithms, perform intelligent searches in the massive space of solutions considering the required

**Table 2.** Possible firing patterns of rats' brains single neurons for the BLA and the hippocampus areas.

| Region \ Firing Pattern | Tonic spiking | Phasic spiking | Tonic bursting | Phasic bursting | Mixed mode | Spike freq. ad. | Class 1 excitable | Class 2 excitable | Spike latency | Subthreshold osi. | Resonator | Integrator | Rebound spike | Rebound burst | Threshold Var. Variability | Bistability | DAP | Accommodation | Inhibition spike. | Inhibition burst. |
|---|---|---|---|---|---|---|---|---|---|---|---|---|---|---|---|---|---|---|---|---|
| BLA | + | + | – | – | – | – | – | – | – | – | + | + | – | + | – | + | – | + | – |
| Hippocampus | + | – | – | – | + | – | – | – | – | – | – | – | – | – | – | – | – | – | + | – |

**NEUROSCIENCE**

statistical techniques. The natural approach followed by these algorithms results in achieving optimal solutions for natural phenomena, like neurons' spiking activity. Consequently, one of the best optimization algorithms, a genetic algorithm, was used in this inquiry.

To desirably represent the effectiveness of genetic approach in optimizing the Izhikevich model parameters, we compared the firing pattern of model corresponded to optimized parameters with the firing pattern related to real data recorded from the BLA and HIP neurons. Modeling results have indicated that the rat real neurons activity and the improved Izhikevich pattern have a desirable adaptation.

The value of Izhikevich model parameters, before and after applying optimization is represented in Table 1. In all cases, one substantial point related to optimized parameters is that the values of parameters and have been reinforced with a large rate. Other parameters have changed with a relatively slow rate, based on their spiking behavior. A larger amount of parameter, i.e. the after-spike reset of the variable, suggests a larger after-spike jump of variable in the behavior of the real single neurons. Additionally, larger values of parameter result in the faster recovery of variable . Greater values of the parameter represent stronger subthreshold fluctuations in neurons firing pattern, according to the values of the variables and . The parameter has changed in the range of -65 to -50. Larger amounts of this parameter result in deep voltage reset. In conclusion, in vivo electrophysiological data recorded from the rat HIP and BLA represented larger after-spike jump of variable, faster recovery of variable, increasing or diminishing in low-threshold spiking dynamics, and deeper or shallower voltage reset, compared to the original Izhikevich neuron.

As an important result of this study, in the future representation of the firing activity of rat HIP and BLA neurons under normal activity, this improved model could be applied concerning the optimized values of Izhikevich neuronal model parameters. This could result in a great simplification in the simulation of large-scale neural networks and the development of their hidden layers.

More importantly, our research was the first study investigating the possible firing patterns following the rat HIP and BLA neurons under normal activity and anesthesia, from a mathematical point of view. The capability to assign a distinctive parametric model to each potential following firing pattern has a great implication in computational neuroscience and is a support for the concept that the form of distinctive firing patterns may influence representing the neuron's function. Additionally, future network modeling of spiking neurons based on the rat HIP and BLA neuronal activity can be performed by only inserting the possible firing patterns and excluding the cases that cannot be followed by the single neurons of these brain areas. The potential firing patterns of the BLA and HIP single neurons are represented in Table 2.

Some of the irrelevant patterns, i.e. not following by the HIP and BLA neurons are shown in Figure 2 C-E, as well as Figure 3 E and F. In addition to these figures, there are also other firing patterns that are not compatible with the normal neural activity of the HIP and BLA under anesthesia. Potential reasons for the observed irrelevance are differences in spike timings, the number of spikes in a specific time period, depolarization and repolarization timing and shape, either with or without applying improvement to the conducted model. We avoided further explanations regarded those patterns in this study.

To support the validity of the achieved results, data recording was performed for several rats under anesthesia, and the whole analysis processes were repeated for acquired data. The firing activity of all BLA and HIP neurons was compared; for the recorded data from each region, the obtained results were predominantly consistent. Future research studies may benefit from recording data from the BLA and HIP under the effect of drugs or in awake animals. They are suggested to investigate whether the Izhikevich model or our modified model is capable of representing those neural activities. Moreover, it is noteworthy to explore the changes in possible firing patterns following neurons under the effect of a special drug or awaking in comparison to the normal condition under anesthesia. Another interesting future study can be the investigation of other optimization algorithms and compare their effectiveness in improving the Izhikevich model to represent the neural activity of different rat brain areas efficiently.

5. Conclusion

This study was conducted to elevate our knowledge of neural interactions in different structures of the brain and accelerate the quality of future large-scale neural networks simulations, as well as reducing the modeling complexity. This aim was achievable by performing the improved Izhikevich model, and inserting only the plausible firing patterns and eliminating unrealistic ones.

## Ethical Considerations

### Compliance with ethical guidelines

All experiments were performed following the National Institutes of Health Guide for the Care and Use of Laboratory Animals (NIH Publication No. 80-23, revised 1996) and were approved by the Research and Ethics Committee of Shahid Beheshti University of Medical Sciences, Tehran, Iran.

### Funding

This work received no financial support.

### Authors' contributions

All authors contributed in preparing this paper.

### Conflict of interest

The manuscript has not been previously published, is not currently submitted for review to any other journal, and will not be submitted elsewhere before a decision is made by this journal. The authors declared no conflicts of interest.